\documentclass[twoside,11pt]{article}

\usepackage{amssymb}
\usepackage{amsmath}
\usepackage{graphicx}
\usepackage{url}
\newcommand{\xiv}{\mathbf{\xi}}
\newcommand{\xv}{\mathbf{x}}
\newcommand{\Fv}{\mathbf{F}}

\newcommand{\vim}{\textsc{vim}}
\newcommand{\oob}{\textsc{oob}}

\newcommand{\Iri}{\textsc{Iri}}
\newcommand{\Irii}{\textsc{Iri2}}
\newcommand{\Mad}{\textsc{Mad}}
\newcommand{\Rnd}{\textsc{Rnd}}
\newcommand{\Rndd}{\textsc{Rnd2}}
\newcommand{\Rnddd}{\textsc{Rnd3}}

\usepackage[utf8]{inputenc}

\begin{document}

\title{Embedded all relevant feature selection with Random Ferns}
\author{Miron Bartosz Kursa  \\
        Interdisciplinary Centre for Mathematical and Computational Modelling\\
       University of Warsaw\\
       Pawińskiego 5A, 02-106 Warsaw, Poland}

\maketitle

\begin{abstract}%
Many machine learning methods can produce variable importance scores expressing the usability of each feature in context of the produced model; those scores on their own are yet not sufficient to generate feature selection, especially when an all relevant selection is required.
Although there are wrapper methods aiming to solve this problem, they introduce a substantial increase in the required computational effort.

In this paper I investigate an idea of incorporating all relevant selection within the training process by producing importance for implicitly generated shadows, attributes irrelevant by design.
I propose and evaluate such a method in context of random ferns classifier.
Experiment results confirm the effectiveness of such approach, although show that fully stochastic nature of random ferns limits its applicability either to small dimensions or as a part of a broader feature selection procedure.
\end{abstract}

%\begin{keywords}
% Feature Importance, Feature Selection, Random Forest, Random Ferns
%\end{keywords}

\section{Introduction}
The crucial part of any machine learning application is to find a good representation of the observed data; its ability to express the contained information in a way that is learnable for the utilised modelling method is often the most important component of the final performance.
On the other hand the robustness of this process is also critical, as it is very easy to introduce overfitting this way; either by leaking information which model should not be given or by amplifying false, random associations which are only present in training data due to its finite size.

The are also cases when the original data is already in a `tabular' form, i.e. as a series of independently measured features, and it is strongly desired to retain this structure because of known links between features and certain physical aspects of the investigated phenomenon.
Then, the only possible form of representation altering is \textit{feature selection}, reducing the original set of attributes to a relevant, sometimes also non-redundant subset.
Such selection may be less effective in terms of learnability, yet becomes an useful result on its own by highlighting important aspects of the problem.

Most prominent examples here are data sets obtained via high throughput biological experiments: they can simultaneously capture the activity of thousands or even millions of agents representing a substantial fraction of a full state of a given system, though only a handful is expected to be connected to the investigated state or behaviour.
Finding important yet previously unknown agents in such case can lead (through targeted studies) to a discovery of novel mechanisms, consequently be more important than the original task of building a black-box model.

Fundamentally, feature selection methods can be divided into two classes, \textit{minimal optimal} and \textit{all relevant}  \cite{Nilsson2007} --- methods of the first group attempt to find a smallest subset of features on which certain model achieves optimal performance; second group collects methods which attempt to remove features irrelevant to the problem, consequently retaining those features which may be useful for modelling.
The first aim is straightforward to implement as an accuracy optimisation problem; unfortunately, such selection is susceptible to contain attributes forming false associations, on the other hand missing true, interesting interactions due to their apparent redundancy.
Consequently, only all relevant selection is applicable for deciphering mechanisms behind the analysed data; also it leads to a better robustness of the selection, especially in a $p\gg n$ setting.

There is also a technical taxonomy due to how the selection is coupled with modelling \cite{Saeys2007}: there are \textit{filters}, algorithms which are independent from the modelling method, \textit{embedded methods} which integrate selection and modelling into a single algorithm, finally \textit{wrappers} which relay on some modelling method, but only as an oracle of an efficiency of a given subset of features or feature importance.
Obviously all filters have to be all relevant methods, yet they often only analyse univariate or very simple interactions between attributes and decision, leading to poor results \cite{Saeys2008}.
There are wrappers belonging to each of fundamental classes, though the universal problem of those methods is their computational demand, as they usually require building hundreds of models.
Embedded methods are generally much faster, though they are usually either minimal optimal or only offer variable importance scores (\vim{}) or ranking rather than strict selection.

All relevant wrappers \cite{Tuv2006,HuynhThu2008,Kursa2010a} generally relay on external feature importance provider, yet try to establish a \vim{} threshold separating relevant and irrelevant features.
While it is mostly impossible to analytically obtain the distribution of \vim{} for irrelevant variables, aforementioned wrappers use a kind of permutation test approach to estimate it --- utilising either permutation of the decision attribute or features irrelevant by design (called \textit{shadows} or \textit{contrasts}) injected within original ones.
Such an approach obviously requires many repetitions to stabilise the approximation; some methods also try to progressively eliminate irrelevant features, as both stability and the quality of \vim{} often decrease with the dimensionality of a set.

The aim of this work is to propose a methodology to implement such all relevant selection in an embedded method setting, bringing together selection quality and computational efficiency.
In particular, I will closely follow the heuristics behind the Boruta method \cite{Kursa2010b}, as it is generic and proved effective in a demanding assessment \cite{Kursa2014a}.
While it can be in principle generalised for any method producing variable importance or ranking, here I will introduce and evaluate it in context of random ferns, an efficient, stochastic ensemble classifier.

\section{Boruta}

Boruta \cite{Kursa2010a,Kursa2010b} is an all relevant feature selection wrapper, originally developed for Random Forest \cite{Breiman2001}, but in a present form capable of using arbitrary classifier that outputs \vim{}.
The algorithm works in a following way.
First, \vim{} is calculated on a modified data set extended with explicitly generated shadows, i.e. nonsense features created from the original ones by permuting the order of values within them (thus wiping out information but preserving distribution).
Then, all original variables which importance was larger than the maximal importance of a shadow are assigned a \textit{hit}.

This procedure is repeated (with shadows re-generated each time), while the proportion of hits for each feature is observed for being significantly lower or higher than half; in the first case the feature is claimed rejected and removed from further consideration, in the latter claimed confirmed and added to final result.
The algorithm stops either when the status of all original features is decided or when a previously given maximal number of iterations is exhausted, in which case some variables may be left undecided.

\section{Random Ferns}
Random ferns are classifiers introduced in \cite{Ozuysal2007} and named as such in \cite{Ozuysal2008}.
They were developed to serve as an computationally efficient alternative to Random Forests in demanding computational vision tasks \cite{Bosch2007,Oshin2009}.
I have previously shown, however, that with certain modifications they can perform well in a generic machine learning context \cite{Kursa2014b}, as well as produce variable importance measure (\vim) of a similar quality as Random Forest, though in a much shorter time \cite{Kursa2014a}.
Here I will briefly present the method in the aforementioned generalised version, called \textit{rFerns}.

Let's assume a set of $N$ training objects $\mathcal{T}\subset\mathcal{X}$, where $\mathcal{X}$ is composed of $M$ attributes, i.e., $\mathcal{X}=\prod_{j=1}^M\mathcal{X}_j$, where $\mathcal{X}_j$ is a domain of the $j$-th attribute.
Each object $\xv\in\mathcal{T}$ is assigned $Y(\xv)$, one of $C$ disjoint classes.

rFerns is an ensemble of $K$ ferns $\Fv^k(\mathcal{T}'):\mathcal{X}\rightarrow \mathbb{R}^C$ returning a vector of class scores indicating confidence how a given object fits within each class.
The ensemble is built using bagging, thus $k$-th fern is built on a separate random multiset of training objects called \textit{bag}, $\mathcal{B}_k:=B(\mathcal{T},k)$, where $B(\mathcal{X},\lambda)$ denotes sampling with replacement the same number of elements as in the input set, and $\lambda$ functions as a random seed of sampling procedure, i.e., samples with different $\lambda$ would be different and statistically independent.
The prediction of the ensemble is the class which gets maximal sum of scores over all ferns, i.e.,
\begin{equation}
 \hat{Y}(\xv):=\arg\max_y\sum_k F^k_{y}(\xv;\mathcal{B}_k).
\end{equation}

The base of $\Fv$ is a \textit{trunk} function $T:\mathcal{X}\rightarrow 1..2^D$ classifying given object into its corresponding \textit{leaf}, which is then associated with a vector of classes' scores estimated using $\mathcal{T}$.
$D$ is a hyperparameter of the classifier and is called \textit{fern depth}.
Trunk is defined by a sequence of $D$ split attribute indexes $j_i\in 1..M$ and subsets $\Xi_i\subset \mathcal{X}_{j_i}$, so that an element $\xv=(x_1,x_2,\ldots,x_M)\in\mathcal{X}$ belongs to a leaf
\begin{equation}
T(\xv):=\sum_{i=0}^D 2^i\cdot I(x_{j_i}\in\Xi_i),
\end{equation}
where $I$ is an indicator function.
As rFerns is a highly stochastic method, trunks are generated randomly: $D$ split attributes are drawn uniformly with replacement from $1..M$, and subsets $\Xi_i$ are either random subsets of $\mathcal{X}_{j_i}$ (for finite, unordered domains) or generated from some random threshold value $\theta_i$, i.e. $\Xi_{i}=\{x\in \mathcal{X}_{j_i}:x\ge \theta_i\}$.

On the other hand, scores are based on the distribution of training objects' classes over leaves, adjusted for the class imbalance and with add-one smoothing.
For convenience, let's first define
\begin{equation}
 \mathcal{L}_{\mathcal{B},T}(\xv)=\left\{\xiv \in \mathcal{B} : T(\xiv)=T(\xv) \right\},
\end{equation}
a subset of bag objects in the same leaf as $\xv$ and
\begin{equation}
 \mathcal{Y}_\mathcal{B}(y)=\left\{\xiv \in \mathcal{B} : Y(\xiv)=y \right\},
\end{equation}
a subset of bag objects of a class $y$.
Full fern is then defined as
\begin{equation}
 \exp(F_{y}(\xv;\mathcal{B})):=
 \frac{1+\#(\mathcal{L}_\mathcal{B}(\xv)\cap \mathcal{Y}_\mathcal{B}(y))}
  {C+\#\mathcal{L}_\mathcal{B}(\xv)}\cdot
 \frac{C+\# \mathcal{B}}{1+\# \mathcal{Y}_\mathcal{B}(y)},
\end{equation}
where $\#$ denotes multiset cardinality.
Note that the score expresses the deviation of class proportion within leaf with respect to what is expected from random assignment; under-represented classes have negative scores while over-represented positive.
For balanced classes and $\mathcal{L}_\mathcal{B}(\xv)=\varnothing$, the score is $\mathbf{0}$.

\section{Shadow Importance}
Similarly to Random Forest, the regular rFerns \vim{} is defined using the \textit{out-of-bag} or \oob{} objects, i.e., those which are not in the bag of a given fern; I denote this set here as $\mathcal{B}^\ast:=\mathcal{T}-\mathcal{B}$.
rFerns \vim{} for an attribute $a$ is defined as
\begin{equation}
I_a=\sum_{k\in \mathcal{A}(a)}\frac{1}{{\#\mathcal{A}(a)\cdot \#\mathcal{B}_{k}^{\ast}}}
\left(
\sum_{\xiv\in \mathcal{B}_{k}^{\ast}}
 F^k_{Y(\xiv)}(\xiv;\mathcal{B}_k)-
\sum_{\xiv^\diamond\in \mathcal{B}_{k,a}^{\ast\diamond}}
 F^k_{Y(\xiv^\diamond)}(\xiv^\diamond;\mathcal{B}_k)
\right),
\label{eq:regvimng}
\end{equation}
where $\mathcal{A}(a)$ is a set of ferns that incorporate feature $a$, and $\mathcal{B}_{k,a}^{\ast\diamond}$ is $\mathcal{B}^\ast_{k}$ in which values within attribute $a$ has been randomly shuffled, thus decoupled from other attributes' values and object classes, consequently making $a$ irrelevant.
The involved permutation is different for every $(k,a)$ pair.

In the Boruta method, distribution of \vim{} of irrelevant attributes is estimated by calculating \vim{} on the training set augmented with shadows, randomly permuted copies of all true features.
This method is straightforward but inefficient --- augmented set occupies twice as much memory, also the search space of potential interactions becomes significantly larger.
To this end the shadow attributes for rFerns shadow \vim{} are created implicitly during the importance calculation loop, utilising fern trunks built only on actual features.
This yields \textit{shadow importance}, which is defined as
\begin{equation}
 \begin{split}
J_a=
\sum_{k\in \mathcal{A}(a)}\frac{1}{{\#\mathcal{A}(a)\cdot \#\mathcal{B}_{k}^{\ast}}}
\sum_{\xiv\in \mathcal{B}_{k}^{\ast}}
 F^k_{Y(\xiv)}(\xiv;B(\mathcal{T}^\diamond_a,k))+\\-
 \sum_{k\in \mathcal{A}(a)}\frac{1}{{\#\mathcal{A}(a)\cdot \#\mathcal{B}_{k}^{\ast}}}
\sum_{\xiv^\diamond\in \mathcal{B}_{k,a}^{\ast\diamond}}
 F^k_{Y(\xiv^\diamond)}(\xiv^\diamond;B(\mathcal{T}^\diamond_a,k)),
\end{split}
\label{eq:shavim}
\end{equation}
where $\mathcal{T}_{a}^{\diamond}$ is $\mathcal{T}$ in which values of an attribute $a$ have been randomly shuffled.
Note that while $\mathcal{B}_{k,a}^{\ast\diamond}$ is different for each $(k,a)$ pair, $\mathcal{T}_{a}^{\diamond}$ is the same for all ferns; this is crucial because implicit shadows are required to mimic the behaviour of original attributes, which obviously do not change due to being used by different ferns.

To implement feature selection with shadow importance, each original attribute importance is compared with the maximum shadow importance, i.e. the selected subset of features is
\begin{equation}
 \mathcal{S}=\{a\in1..M:I_a>\max_{a'}J_{a'}\};
 \label{eq:sel}
\end{equation}
this approach also follows the heuristics behind the Boruta method.

\section{Numerical Experiments}
The proposed feature selection system is assessed on a series of $6$ synthetic problems built based on an established benchmark datasets.
All of them belong to $p\gg n$ class, and contain \textit{a priori} known set of relevant attributes; in case of three of them this set is empty, as they were derived through total randomisation of a benchmark data.
Precisely:
\begin{itemize}
 \item \Iri{} --- a derivative of the \cite{Fisher1936} iris data, enlarged by adding 4996 irrelevant features generated by shuffling original ones.
 \item \Irii{} --- similar to \Iri{}, but with 1000 irrelevant features only.
 \item \Mad{} --- the Madelon dataset from the \textsc{Nips} 2003 feature selection challenge \cite{Nips2003}, a $5$-dimentional \textsc{xor} problem extended with $15$ random linear combinations of $5$ main attributes (also considered relevant in this work) and $480$ irrelevant features containing random values.
 \item \Rnd{} --- a derivative of the \cite{Golub1999} microarray data, made nonsense by randomly shuffling values within each feature.
 \item \Rndd{} --- a derivative of the \cite{Singh2002} microarray data, made nonsense in a same way as in case of \Rnd{}.
 \item \Rnddd{} --- another nonsense derivative of the Singh et al. data, however made by randomly shuffling values within decision (so that all inter-attribute relation present in the original data are retained).
\end{itemize}

For each of those datasets, a series of rFerns-based embedded feature selections has been performed: over $10$ repetitions with different random seeds and over a comprehensive subspace of hyperparameters.
The investigated values of the depth parameter $D$ were $1$, $3$, $5$, $7$, $9$, $10$ and $12$.
For sake of comparability between cases for different $D$ and sizes of the data, the ensemble size parameter $K$ was set to value corresponding to a given average number of considerations (\textit{scans}) of a feature, i.e. $\left<\#\mathcal{A}(a)\right>_a$; here the scan has been performed for values in range $10$--$5000$.

As a base for comparison, a similar analysis has been performed with Boruta using, as in its original implementation, Random Forest \vim{}; it was also stabilised by taking 10 repetitions and applied for $500$, $5000$ and $50000$ trees in ensemble.
This is because Random Forest has no parameter corresponding to fern depth, while the number of scans is not uniform across features due to a greedy generation of trees.
To compare the results of Boruta and ferns, the average, single core running time of both algorithms was used as an indication of the profoundness of a method in a place of the average attribute scan count.

\begin{figure}
 \includegraphics[width=\textwidth]{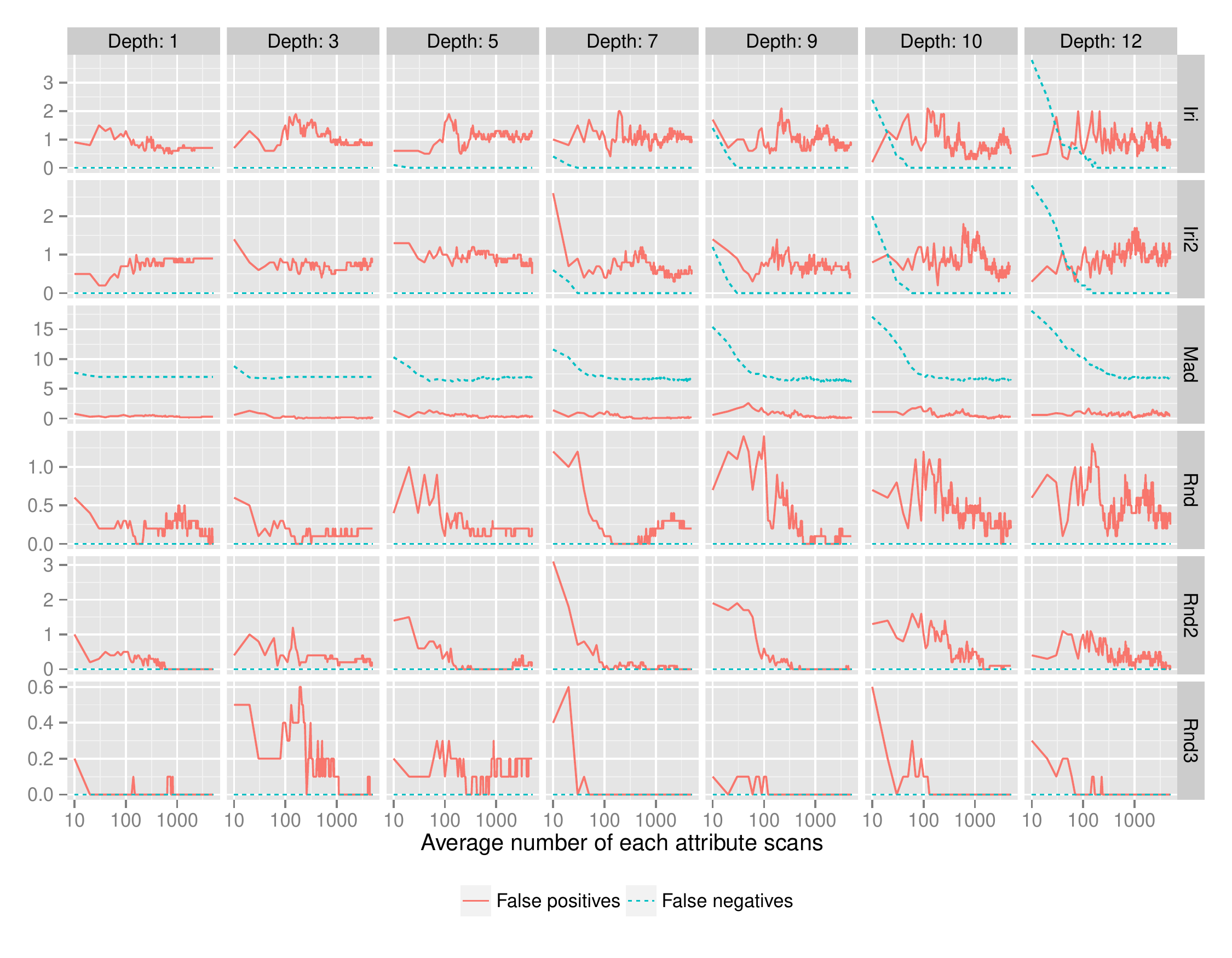}
 \caption{\label{fig:basic} The accuracy of rFerns all relevant feature selection for benchmark problems, shown as the number of false positive and false negative attribute selections.
 Position on the $x$ axis corresponds to to the average number of times each attribute was considered in importance calculation (which is approximately $K/D$); columns correspond to various fern depths while rows to problems.
 Note that for \textsc{Rnd*} problems the number of false negatives is strictly $0$ by design.}
\end{figure}
\begin{figure}
 \includegraphics[width=\textwidth]{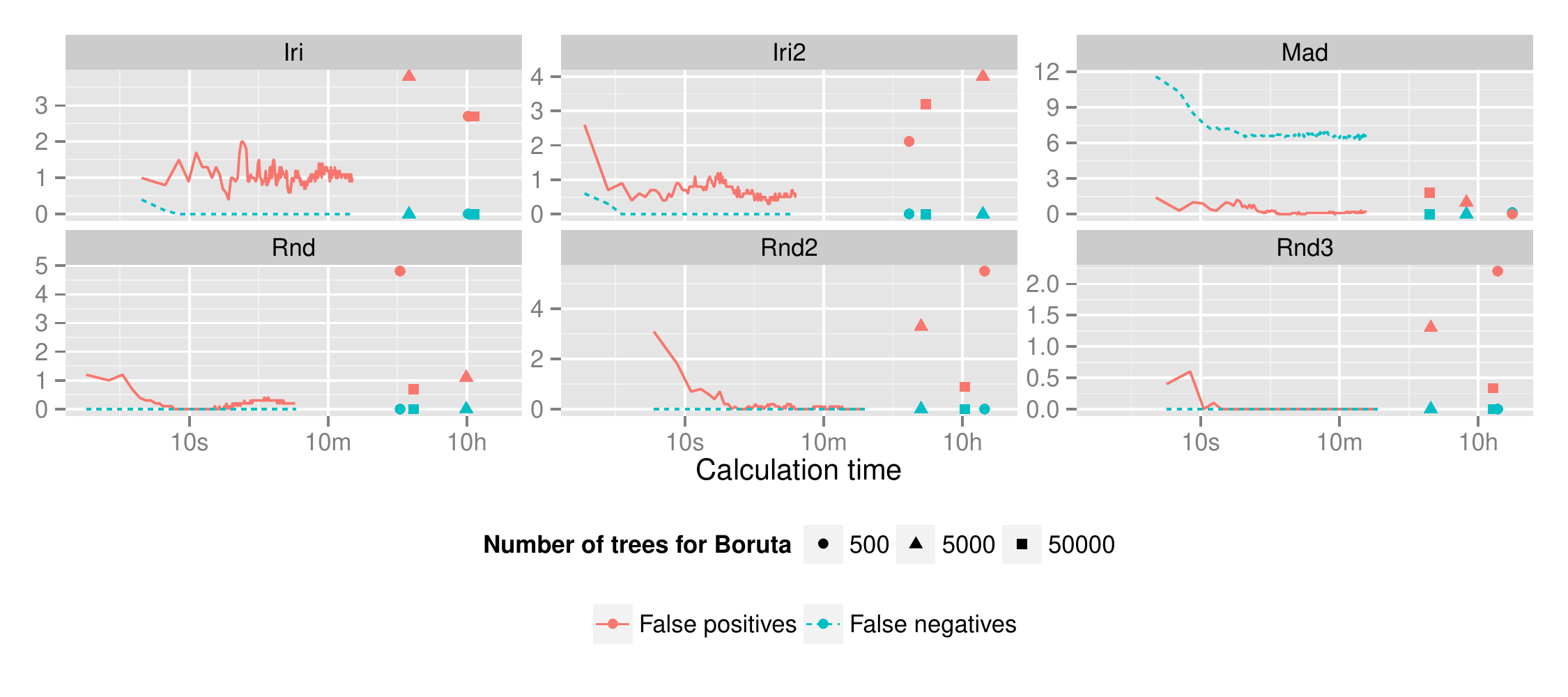}
 \caption{\label{fig:borcomp} Data as on Figure~\ref{fig:basic}, yet shown only for $D=7$, with respect to a single core calculation time and compared with the results of a Boruta feature selection (points).}
\end{figure}

The measured accuracy of rFerns-based selections is presented on Figure~\ref{fig:basic}, and compared with the Boruta result on Figure~\ref{fig:borcomp}.
One can see that in all cases except \Mad{} the selection is very accurate; the number of false negatives always converge to $0$, while the number of false positives is of an order of several attributes, which corresponds to false positive rates of and order of $10^{-4}$.
The quality of the selection is also robust across values of $D$, although deeper ferns require more attribute scans (consequently ferns in ensemble) to converge; overall the selection is most effective for an intermediate value of $D=7$.

The Boruta method yields more accurate result only in case of the \Mad{} set, and generates few times more false positives than rFerns selection for all sets.
When considering computational time, rFerns always managed to converge withing 10 minutes, while Boruta took at minimum 1 hour and 9.4 hours at average\footnote{One should note that those are single-core timings, while both Random Forest used by Boruta and rFerns are trivially parallel and scale well even up to hundreds of processing cores.}.
It is also worth noticing that the selection error in case of \Rnddd{} (in which all attributes are noise) is substantially smaller than in case of \Rndd{} (in which only the decision is permuted).

\begin{figure}
 \includegraphics[width=\textwidth]{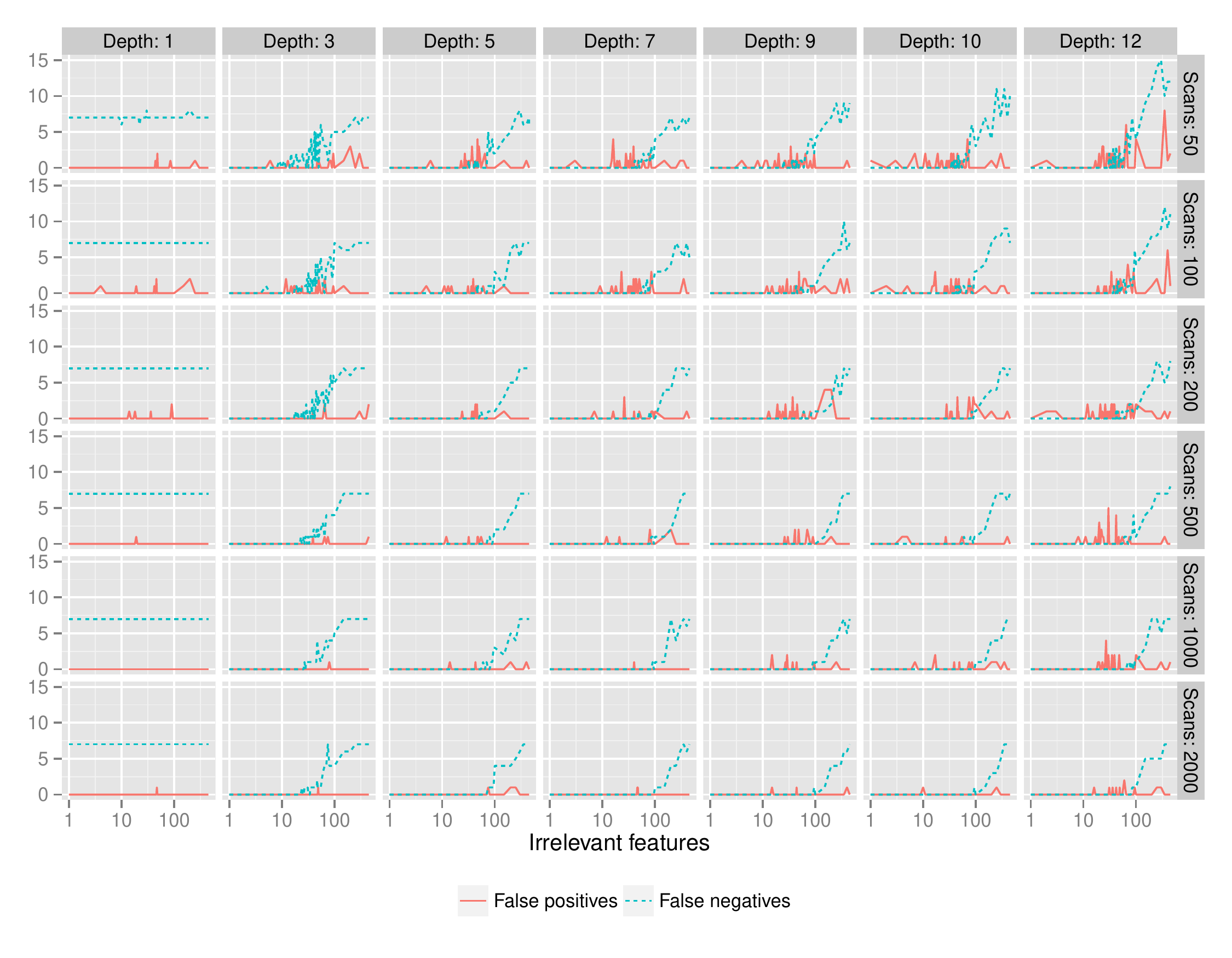}
 \caption{\label{fig:mad} The results of rFerns all relevant feature selection for a family of Madelon dataset modifications made by trimming the original set of $480$ irrelevant attributes to a random subset of from $1$ to $480$ of them (corresponding to the position on the $x$ axis).
 The plot shows the number of false positive and false negative attribute selections; columns correspond to various fern depths, while rows to the average number of times each attribute was considered in importance calculation (which is approximately $K/D$).}
\end{figure}

Because of an inferior performance of rFerns with comparison to Boruta on the \Mad{} data, I decided to study this problem deeper, namely how the selection accuracy depends on the number of nonsense features in the training set.
To this end, the original data was spread into a family of derived problems \Mad$_w$ of a controlled complexity --- namely, \Mad$_w$ contains only $w$ of $480$ irrelevant features present in the original Madelon dataset.
This test was performed by applying rFerns feature selection on \Mad$_w$ for $w\in\{1,2,3,\ldots,50,55,$ $60,\ldots,150,200,250,\ldots,450\}$ and, similarly to previous experiments, using $10$ repetitions, for $D\in\{1,3,5,7,9,10,12\}$ and for the average number of scans equal to $50$, $100$, $200$, $500$, $1000$ and $2000$.

The results of this analysis are summarised on Figure~\ref{fig:mad}.
One can see that the \textsc{xor} problem underlying \Mad{} data is solvable perfectly by the rFerns method for $D\ge3$, although up to only around $100$ irrelevant features ($5\times$ the number of relevant ones) in case of $D=7$ and more than $1000$ scans of each attribute.
Still, for a reasonable amount of scans the method never misses more than $7$ features; while those are the same attributes that $D=1$ rFerns always miss, it is likely that those are relevant features which can only be identified in multivariate analysis.
Rare false positives seem to arise at random, with probability rising with $D$ and the amount of noise in the system; though they can be practically eliminated by increasing the number of scans.

Those results suggests that the poor performance of rFerns selection on the full \Mad{} problem is caused by the fact that trunks are built stochastically and not optimised, as in Random Forest utilised by Boruta, thus is not fundamental to the construction of the feature selection methodology.

\section{Conclusions}
The paper proposes an approach to extend basic variable importance score into an embedded, all relevant feature selection method.
It is based on the construction of the Boruta wrapper method, in particular the idea of shadow attributes; they are though generated only implicitly, which leads to a better efficiency and smaller memory footprint.

The results of numerical assessment of the obtained method show that it is highly specific, robust and computationally efficient.
On a range of problems it can par or even outperform Boruta, still achieving substantial speed-ups.
Due to a fully stochastic nature of the underlying rFerns classifier, its capability of detecting complex multivariate interactions is though susceptible to a high dimensionality of the input set.
Consequently, it seems that applying rFerns feature selection for complex problems with hundreds or more of features would require adding some amount of optimisation; for instance, trunk building may be biased to include important features more often, or the entire method wrapped in a forward selection scheme.

% The implementation of described algorithms are implemented in the version 2.0.0 of the rFerns software, available at \url{https://github.com/mbq/rFerns/tree/9da46d}.

\section*{Acknowledgements}
{\footnotesize
This work has been financed by the National Science Centre, grant 2011/01/N/ST6/07035, as well as with the support of the \textit{\textsc{OCEAN} --- Open Centre for Data and Data Analysis} Project, co-financed by the European Regional Development Fund under the Innovative Economy Operational Programme.

\noindent Computations were performed at the ICM, grant G48-6.}

\bibliographystyle{alpha}

\bibliography{text}{}

\end{document}